\documentclass[runningheads]{llncs}

\usepackage{pifont}
\newcommand{\cmark}{\text{\ding{51}}}
\newcommand{\xmark}{\text{\ding{55}}}
\usepackage{amssymb, amsmath, amsfonts}
\usepackage{euscript}
\usepackage{mathtools}
\usepackage{hyperref}
\usepackage{cleveref}
\usepackage{subcaption}
\usepackage{graphicx}
\usepackage{caption}
\usepackage{booktabs}

\usepackage{comment}
\usepackage{graphicx}
\usepackage[table]{xcolor}
\usepackage{hyperref}
\graphicspath{Figures/}
\usepackage{multirow}

\hypersetup{
    colorlinks=true,
    linkcolor=blue,
    filecolor=cyan,      
    urlcolor=magenta,
    }
\begin{document}

\title{Self-supervised Few-shot Learning for Semantic Segmentation: An Annotation-free Approach}

\titlerunning{Annotation-free}

\author{Sanaz Karimijafarbigloo\inst{1} \and
Reza Azad\inst{2} \and
Dorit Merhof\inst{1,3}}

\authorrunning{S. Karimijafarbigloo et al.}

\institute{Institute of Image Analysis and Computer Vision, Faculty of Informatics and Data Science, University of Regensburg, Germany \and
Faculty of Electrical Engineering and Information Technology, RWTH Aachen University, Aachen, Germany \and
Fraunhofer Institute for Digital Medicine MEVIS, Bremen, Germany\
\email{dorit.merhof@ur.de}}

\maketitle              

\begin{abstract}
Few-shot semantic segmentation (FSS) offers immense potential in the field of medical image analysis, enabling accurate object segmentation with limited training data. However, existing FSS techniques heavily rely on annotated semantic classes, rendering them unsuitable for medical images due to the scarcity of annotations. To address this challenge, multiple contributions are proposed: First, inspired by spectral decomposition methods, the problem of image decomposition is reframed as a graph partitioning task. The eigenvectors of the Laplacian matrix, derived from the feature affinity matrix of self-supervised networks, are analyzed to estimate the distribution of the objects of interest from the support images. 
Secondly, we propose a novel self-supervised FSS framework that does not rely on any annotation. Instead, it adaptively estimates the query mask by leveraging the eigenvectors obtained from the support images. This approach eliminates the need for manual annotation, making it particularly suitable for medical images with limited annotated data. Thirdly, to further enhance the decoding of the query image based on the information provided by the support image, we introduce a multi-scale large kernel attention module. By selectively emphasizing relevant features and details, this module improves the segmentation process and contributes to better object delineation.
Evaluations on both natural and medical image datasets demonstrate the efficiency and effectiveness of our method. Moreover, the proposed approach is characterized by its generality and model-agnostic nature, allowing for seamless integration with various deep architectures. The code is publicly available at \href{https://github.com/mindflow-institue/annotation_free_fewshot}{\textcolor{magenta}{GitHub}}.

\end{abstract}


\section{Introduction}
Computer vision tasks such as localization and segmentation, which require a detailed understanding of image structure, can achieve good results when approached with fully-supervised deep learning methods. Although the success of supervised deep learning methods depends heavily on the availability of a large amount of well-annotated data~\cite{chaitanya2020contrastive,azad2023advances}, collecting and annotating such data is costly and challenging, as it requires to be performed manually by a domain expert. The other equally problematic challenge with fully-supervised models is their inflexibility when confronted with new classes of segmentation targets (e.g., different and novel lesion types)~\cite{ouyang2022self,karimijafarbigloo2023mmcformer}. This is a significant challenge, as training a new model for every new segmentation class is impractical and time-consuming. To address the aforementioned problems, few-shot semantic segmentation (FSS) has been proposed.
The core concept of FSS is a potent approach that effectively minimizes the requirement for extensive annotation, enabling precise predictions of unobserved classes using only a limited number of guiding examples. By capitalizing on FSS, a model can create a discriminative representation of a previously unknown class using a small set of labeled examples (support). This acquired representation can then be employed to accurately predict the outcomes for unlabeled examples (query), without the need for any model retraining. 
This approach significantly alleviates the annotation burden and empowers the model to swiftly generalize and adapt to new unseen classes (e.g., new lesions)

Several approaches have been proposed to tackle the FSS problem. One approach involves the use of a mask average pooling strategy, which effectively removes irrelevant features based on information from the support masks~\cite{zhang2020sg}. Another improvement proposed by Wang et al.~\cite{wang2019panet} is the introduction of a novel prototype alignment regularization between support and query images, resulting in better generability for new classes. Additionally, in other recent works~\cite{zhang2019canet}, researchers have utilized deep attention mechanisms to learn attention weights between support and query images, enabling improved label propagation. 
In spite of the promising outcomes observed in applying few-shot learning paradigms to the segmentation of natural images~\cite{zhou2023mceenet}, their utilization in medical image segmentation remains limited. This limitation is due to the scarcity of annotated classes, which hinders the network's ability to generalize and to prevent overfitting~\cite{xiao2023siamese}. The concept of few-shot segmentation on medical images was initially introduced by~\cite{mondal2018few}. The authors proposed the use of adversarial learning to segment brain images, leveraging only one or two labeled brain images, drawing inspiration from successful semi-supervised approaches~\cite{souly2017semi}. 
Feyjie et al.~\cite{feyjie2020semi} introduced a novel approach that incorporates a semi-supervised mechanism within the conventional few-shot learning framework. This approach leverages the availability of abundant unlabeled datasets to predict skin lesion masks for previously unseen samples. In recent work, to further benefit from unlabelled data, Ouyang et al.~\cite {ouyang2022self} proposed a self-supervised few-shot semantic segmentation (FSS) framework called SSL-ALPNet to segment medical images by utilizing superpixel-based pseudo-labels as supervision signals. This method also improved the segmentation accuracy using an adaptive local prototype pooling module.
Xiao et al.~\cite{xiao2023siamese} proposed a Siamese few-shot network for medical image segmentation and they used a grid attention module to enhance semantic information localization.
Ding et al.~\cite{ding2023few} designed a self-supervised few-shot network to segment medical images. They introduced a Cycle-Resemblance Attention module to effectively capture the pixel-wise relations between query and support medical images.

Despite the incorporation of semi-supervised and self-supervised techniques within these strategies to optimize the training procedure of the model, the presence of annotated data remains indispensable during the inference stage for accurate query mask prediction. To mitigate this requirement, we undertake an exploration of the role played by self-supervised techniques in facilitating the acquisition of object representation within a conventional few-shot context. Specifically, we draw inspiration from the accomplishments of few-shot segmentation methods in natural images, which rely on the episodic training paradigm. In our approach (depicted in \Cref{{fig:proposed-method}}), \ding{182} we aim to eliminate the need for extensive annotation by leveraging the eigenvectors of the Laplacian matrix derived from the feature affinity matrix of self-supervised networks. This allows us to effectively capture the global representation of the object of interest in the Support image. By integrating this concept into the standard few-shot segmentation framework, \ding{183} we propose an end-to-end network that leverages support guidance to predict the query mask. In order to enhance the decoding process of the query image by leveraging the information from the support image, \ding{184} we propose to incorporate large kernel attention along with multi-scale attention gate modules. These modules effectively highlight pertinent features and intricate details, resulting in an enhanced segmentation process.

\begin{figure}[!th]
    \centering
    \includegraphics[width=1\textwidth]{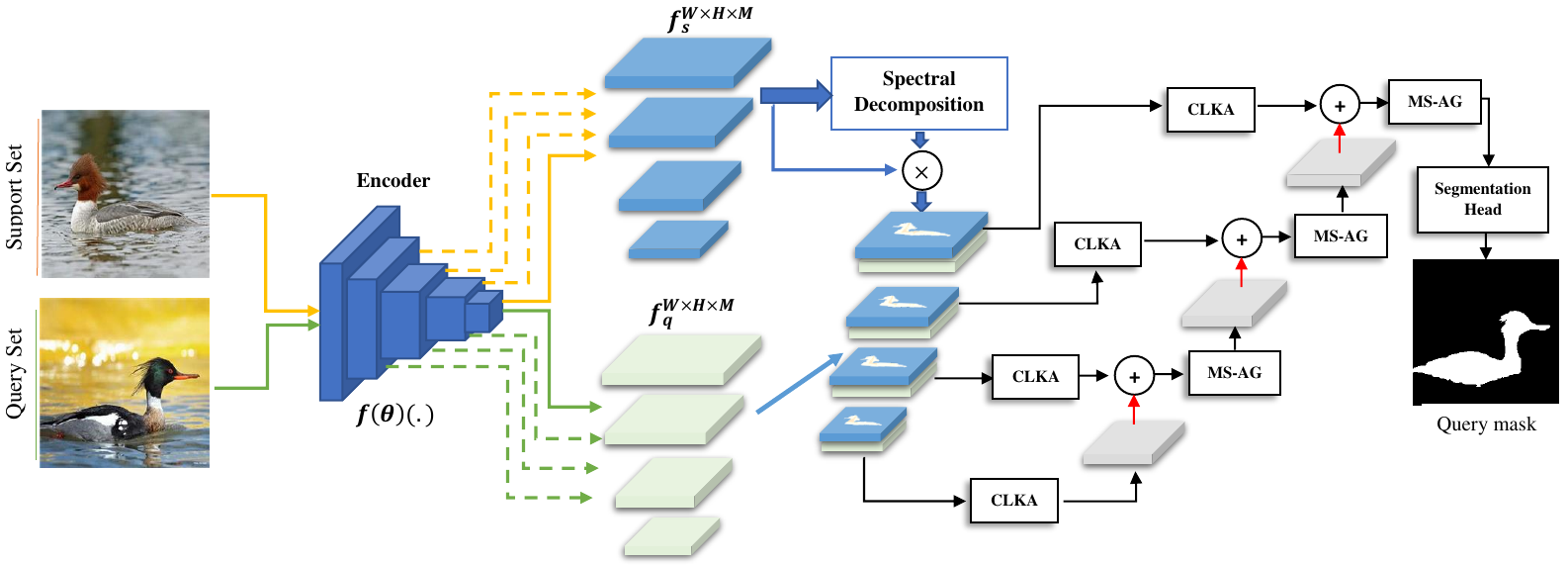}
    \caption{The overview of our annotation-free FSS model.}
    \label{fig:proposed-method}
\end{figure}

\section{Proposed Method} 
\subsection{Problem Formulation}
\label{subsec:problem}
In the context of standard FSS, our approach involves three main datasets: a training set denoted as $D_{train}=\{{(X_i^t,Y_i^t)}\}_{i=1}^{N_{train}}$, a support set denoted as $D_{support}=\{{(X_i^s,Y_i^s)}\}_{i=1}^{N_{support}}$, and a test set denoted as $D_{test}=\{{(X_i^q)}\}_{i=1}^{N_{test}}$. Here, $X_i$ and $Y_i$ represent the input image and corresponding binary mask, respectively. Each dataset contains a total of $N$ images, specified by $N_{train}$, $N_{support}$, and $N_{test}$, involving $C$ distinct classes. Notably, the classes are shared between the support and test sets but are disjoint with the training set, denoted as $\{C_{train}\} \cap \{C_{support}\}=\emptyset$.

The objective of few-shot learning is to train a neural network $f_{(\theta,\gamma)}(\cdot)$ on the training set, enabling it to accurately segment a new class $c \notin C_{train}$ in the test set based on $k$ reference samples from $D_{support}$. Here, $\theta$ and $\gamma$ represent the learnable parameters of the encoder and decoder respectively. 
To reproduce this procedure, training on the base dataset $D_{train}$ follows the episodic learning paradigm introduced in~\cite{vinyals2016matching}, where each episode entails a $c$-way $k$-shot learning task. Specifically, each episode is created by sampling two components. Firstly, we construct a support training set for each class $c$, denoted as $D_{train}^{\mathcal{S}}=\{{(X_s^t,Y_s^t(c))}\}_{s=1}^{k} \subset D{train}$, where $Y_s^t(c)$ represents the binary mask corresponding to the support image $X_s^t$ for class $c$. Secondly, we create a query set $D_{train}^{\mathcal{Q}}=\{{(X_q^t,Y_q^t(c))}\} \subset D_{train}$, where $X_q^t$ is the query image and $Y_q^t(c)$ is the corresponding binary mask for class $c$. In order to estimate the segmentation mask of a given class $c$ in the query image, the model leverages the support training set and the query image. This process can be expressed as $\hat Y_q^t(c)=f_{(\theta,\gamma)}(D_{train}^{\mathcal{S}},X_q^t)$.

More specifically, in our approach we utilize an encoder module to encode the support and query images, resulting in feature representations denoted as $f_s \in \mathbb{R}^{W \times H \times M}$ and $f_q \in \mathbb{R}^{W \times H \times M}$, respectively. Here, $W$, $H$, and $M$ represent the width, height, and feature dimensionality in the feature space, respectively. 
In the subsequent step, we employ a hierarchical approach to acquire the class prototypes, employing a self-supervision strategy in contrast to the prevailing literature~\cite{azad2021texture,hariharan2015hypercolumns}, which utilizes the support mask $Y_s$ to filter out the support prototype. We will provide a full explanation of our hierarchical prototype estimation process in the next sections.

\begin{figure}[!th]
    \centering
    \includegraphics[width=1\textwidth]{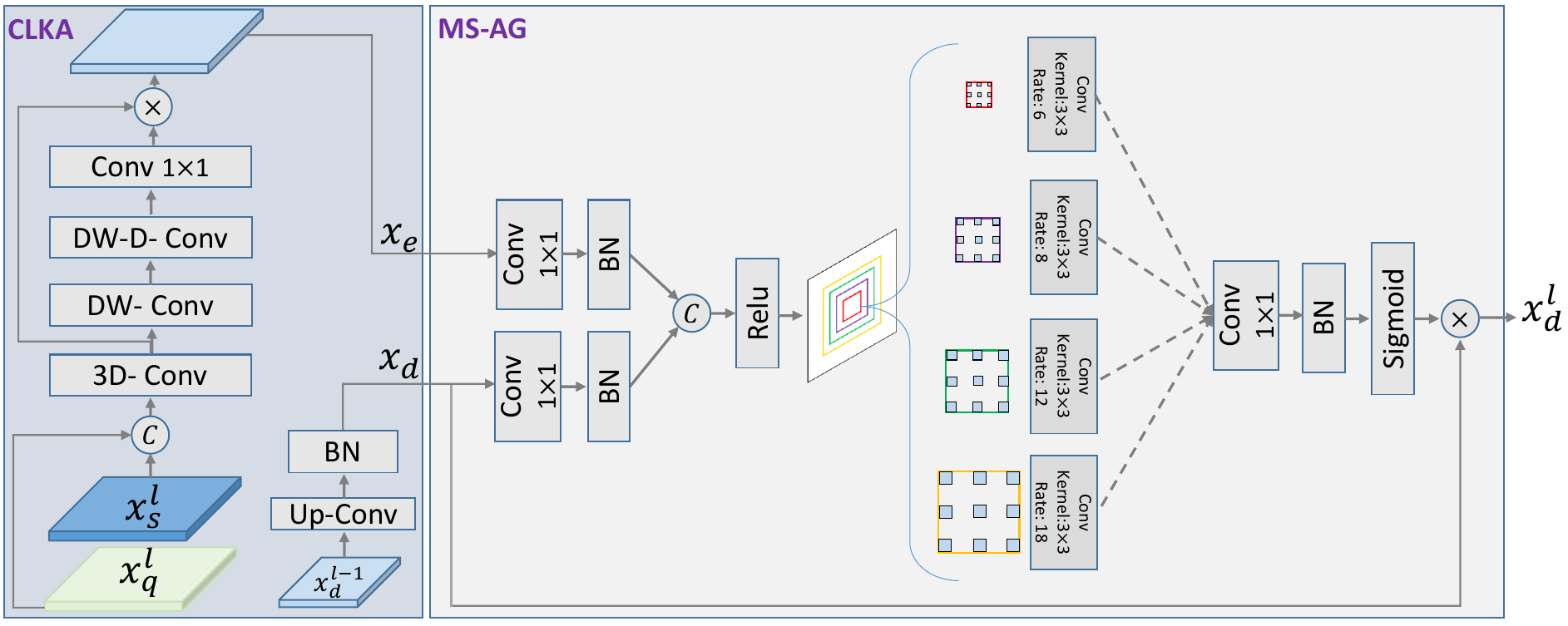}
    \caption{The overview of the proposed \textbf{CLKA} and \textbf{MS-AG} modules. In each block of the decoder network, we include both CLKA and MS-AG for conditioning the query representation based on the support prototype.}
    \label{fig:decoder}
\end{figure}

\subsection{Hierarchical Prototypes}
In the realm of few-shot learning, the support prototype assumes a pivotal role as a representative reference for a specific class, greatly influencing the model's ability to generalize and accurately predict unseen instances. By encapsulating the fundamental characteristics of a class, the support prototype empowers the model with the capacity to make informed predictions. Our study introduces a novel approach for generating a hierarchical support prototype using spectral decomposition, eliminating the need for a support mask.

Initially, we extract the support representation denoted as $f_s \in \mathbb{R}^{W \times H \times M}$ by leveraging an encoder module $f_{(\theta)}(\cdot)$. This support representation is derived from different parts of the encoder module, including combinations of various layers such as hyper-columns~\cite{hariharan2015hypercolumns}. Our experimental findings, consistent with previous research~\cite{azad2021texture}, reveal that incorporating features from both shallow and deep layers of the encoder network produces favorable results. This approach captures multi-level and multi-scale representations while preserving global contextual object features.

Subsequently, we construct an affinity matrix based on pixel correlations. By setting the affinity threshold to zero, our focus lies on aggregating similar features rather than anti-correlated ones. The resulting feature affinities, denoted as $W_\mathrm{s} \in \mathbb{R}^{HW \times HW}$, encompass semantic information at both coarse and low-level resolutions.
Utilizing $W_\mathrm{s}$, we compute the eigenvectors of its Laplacian matrix $L = D^{-1/2}(D - W)D^{-1/2}$ to decompose an image into soft segments, represented as ${y_0, \cdots, y_{n-1}} = \mathrm{eigs}(L)$. Among these eigenvectors, we pay particular attention to the remaining ones $y_{>0}$ since the first eigenvector $y_0$ is constant, corresponding to an eigenvalue $\lambda_0 = 0$.

To identify the support object, akin to prior studies~\cite{melas2022deep}, we examine the Fiedler eigenvector $y_1$ of $L$ and discretize it by considering its sign, resulting in binary image segmentation. By creating a bounding box around the smaller region, which is more likely to represent the foreground object rather than the background, we establish an alternative to the support mask. This bounding box serves the purpose of hierarchically filtering the support representation to generate support prototype $f'_s$.

\subsection{Decoder}
In our network architecture, we incorporate a decoder module consisting of four blocks. Each block follows a specific sequence of operations. Firstly, we employ the cross-LKA (CLKA) module to effectively integrate the query representation with the support prototype. This module aids in capturing meaningful relationships between the query and prototype, enhancing the overall feature fusion process. Subsequently, we utilize the multi-scale attention gate mechanism to combine the output of the CLKA module with the up-sampled features obtained from the previous decoder layer. The multi-scale attention gate (MS-AG) facilitates the selective integration of relevant spatial information from different scales, promoting the refinement of the feature representation. In the next subsections, we will elaborate on cross-LKA and MS-AG in more detail.

\subsubsection{Large Kernel Attention (LKA)}
The attention mechanism, also known as an adaptive selection process, has the ability to identify discriminative features while disregarding noisy responses with respect to the input features. Generating an attention map that signifies the importance of various parts is one of the key roles of the attention mechanism. There are two well-known attention mechanisms and each one has its own pros and cons. The first one is the self-attention mechanism~\cite{dosovitskiy2020image} which has the potential to discover long-range dependencies however it has some drawbacks (e.g., ignoring channel adaptability, high quadratic complexity for high-resolution images, neglecting the 2D structure of images). The second one is large kernel convolution~\cite{park2018bam} which can establish relevance and produce an attention map. Nevertheless, employing large-kernel convolutions introduces substantial computational overhead and increases the number of parameters. To address the mentioned limitations and leverage the advantages of both self-attention and large kernel convolutions, the large kernel attention (LKA) approach is proposed in~\cite{guo2022visual}, which decomposes a large kernel convolution operation to capture long-range relationships.
In our study, we extend the idea of the LKA for distilling discriminative representation derived from the support prototype into a query representation to condition the prediction of the query mask based on support distribution. To this end, we first fuse the support prototype $f'_S$ and the query representation $f_q$ with learnable parameters (modeled as 3D convolution) followed by a non-linear activation function. This operation enables the network to encode the prior knowledge obtained from the support representation into query features for estimating the object of interest's representation and eliminating the background noise. Next, to capture long-range dependency in an efficient way, we follow the LKA approach. Regarding $C$ as the number of channels, then a $C \times C$ convolution can be decomposed into a $[\frac{c}{d}]\times [\frac{c}{d}]$ depth-wise dilation convolution (a spatial long-range convolution) with dilation d, a $(2d-1)\times(2d-1)$ depth-wise convolution (a spatial local convolution) and a $1\times1$ convolution (a channel convolution). Therefore, long-range relationships can be extracted within a feature space and the attention map is generated with a few computational complexity and parameters. The large kernel attention (LKA) module is written as

\begin{equation}
\text { Attention }=\operatorname{Conv}_{1 \times 1}(\mathrm{DW}-\mathrm{D}{-}\operatorname{Conv}(\mathrm{DW}{-}\operatorname{Conv}(\mathrm{{F(f'_s, f_q)}})))
\end{equation}
\begin{equation}
\text { Output }=\text { Attention } \otimes \mathrm{{F(f'_s, f_q)}}
\end{equation}

\noindent where $F(f'_s, f_q) \in {R}^{C \times H \times W}$ and $Attention \in {R}^{C \times H \times W}$ denote the 3D convolutional operation for support and query aggregation and the attention map, respectively. Also, $\otimes$ indicates the element-wise product and the value of the attention map represents the importance of each feature. Unlike conventional attention methods, the LKA approach does not use an additional normalization function such as sigmoid or SoftMax. The overall process is depicted in \Cref{fig:decoder}

\subsubsection{Multi-scale Attention Gate (MS-AG)}
The main purpose of AGs is to mitigate irrelevant features in background regions by employing a grid-attention technique that considers the spatial information of the image~\cite{oktay2018attention}. 
To achieve this objective, we initiate the fusion process by combining the feature representation obtained from the decoder block $x_d^{l-1}$ with the output of the CLKA module $x_e^l$. This fusion is accomplished using a $1{\times}1$ convolution operation, which combines the two sets of information into a unified representation.
Next, to model multi-scale representation we employ Atrous convolution in our attention gate module. Atrous convolution, also referred to as dilated convolution, is a technique that expands the kernel size of a filter without increasing the parameter count or computational load. By introducing $r-1$ zeros between consecutive filter values, the kernel size of a $k \times k$ filter is effectively enlarged to $k_{Atrous} = k + (k-1)(r-1)$. 
Using this multi-scale attention mechanism allows the model to more precisely determine the importance of each region and effectively manage their impact on the final outcome. 
The multi-scale attention gate $MS{-}AG (\cdot)$ can be formulate as follows:

\begin{equation}
q_{a t t}(x_e, x_d)=C_{at}(\sigma_1\left(B N\left(C_e(x_e)+B N\left(C_d(x_d)\right)\right)\right))
\end{equation}
\begin{equation}
MS{-}A G(x_e, x_d)=x_d * \sigma_2\left(B N\left(C\left(q_{a t t}(x_e, x_d)\right)\right)\right)
\end{equation}

\noindent where  $\sigma_1(\cdot)$ refers to ReLU, and $\sigma_2(\cdot)$ corresponds to the Sigmoid activation function. $C_e(\cdot), C_d(\cdot)$, and $C(\cdot)$ indicate the channel-wise $1 \times 1$ convolution operation. $B N(\cdot)$ denotes the batch normalization operation and $C_{at}(\cdot)$ shows the Atrous convolution operation. $x_d$ and $x_e$ represent the up-sampled and skip connection features, respectively. Figure \ref{fig:decoder} illustrates the overall process.

\section{Experiments}
\subsection{Dataset}
In this study, the FSS-1000 dataset is utilized to assess the effectiveness of our method in analyzing natural images. Additionally, to examine the network's ability to generalize to medical images, we evaluate its performance on the publicly accessible ($PH^2$) dataset, specifically designed for skin lesion segmentation. 

\noindent \textbf{FSS-1000:} The FSS-1000 class dataset~\cite{li2020fss} is a significantly large-scale dataset specifically tailored for few-shot segmentation tasks. It encompasses a total of 1000 classes, with each class consisting of 10 images accompanied by their corresponding pixel-level ground truth annotations. The official training split, comprising 760 classes, is utilized as the primary dataset for training purposes. On the other hand, the testing set, comprising 240 classes, is used for inference.

\noindent\textbf{$PH^2$ dataset:} The $PH^2$ dataset~\cite{mendoncca2013ph} consists of 200 RGB dermoscopic images of melanocytic lesions including 80 common nevi, 80 atypical nevi, and 40 melanomas. The dataset was provided at the Dermatology Service of Hospital Pedro Hispano in Matosinhos, Portugal. The resolution of images is 768x560 pixels, but in our work we resized them to $224{\times}224$ pixels. 
In our experimental setup, we follow the same setting suggested in~\cite{feyjie2020semi} to evaluate our method.

\subsection{Implementation Details}
In our implementation, ResNet50 backbone network with ImageNet pre-trained weights is used. Feature extraction is performed by extracting features from the last convolutional layer at each encoder block of the backbone network. This feature extraction approach yields four pyramidal layers ($P=4$).
To ensure consistency, we set the spatial sizes of both support and query images to $400 \times 400$ pixels, resulting in $H,W = 400$. Consequently, we obtain the following spatial sizes for each pyramidal layer: $H_{1},W_{1} = 100$, $H_{2},W_{2} = 50$, $H_{3},W_{3} = 25$, and $H_{4},W_{4} = 13$. As a result, our decoder component consists of four blocks, where each block involves fusing the support prototype with the query representation, as illustrated in \Cref{fig:proposed-method}. The entire network is implemented using the PyTorch framework and optimized using the Adam optimizer, with a learning rate of $1e-3$. To prevent the pre-trained backbone networks from learning class-specific representations from the training data, we freeze the encoder weight.

\subsection{Evaluation Metrics}
For FSS-1000 benchmark, we adopt the mean intersection over union (mIoU) as our evaluation metric.
To assess the performance of our network on the skin dataset, we compare our method against the unsupervised \textit{k}-means clustering method, as well as SOTA self-supervised methods such as DeepCluster~\cite{caron2018deep}, IIC~\cite{ji2019invariant}, and spatial guided self-supervised strategy (SGSCN)~\cite{ahn2021spatial}. Our evaluation methodology follows the guidelines outlined in~\cite{ahn2021spatial}.
To evaluate the efficacy of our network, we employ three evaluation metrics: the Dice similarity coefficient (DSC), the Hammoud distance (HM), and the XOR metric.

\subsection{Results}
\textbf{FSS-1000:} We commence our evaluation of the proposed model on the FSS-1000 dataset, considering two distinct settings. In the first setting, the inference process incorporates the support mask to guide the network. We compare our results with recent few-shot methods, including 
DoG~\cite{azad2021texture}, PFENet~\cite{tian2020prior}, HSNet~\cite{min2021hypercorrelation} and etc. The results for 1-shot and 5-shot scenarios are summarized in \Cref{tab:results2}. Remarkably, our models, set new benchmarks in terms of performance while maintaining a minimal number of learnable parameters. With including the support annotation on the inference time, our 1-shot and 5-shot results exhibit substantial improvements of 15.3\% and 14.9\% in mIoU, respectively, compared to the baseline OSLSM method. Furthermore, compared to the recent SOTA approaches, HSNet~\cite{min2021hypercorrelation} and DAN~\cite{wang2020few}, our strategy achieves promising results.

\begin{figure}[!th]
    \centering
    \includegraphics[width=0.85\textwidth]{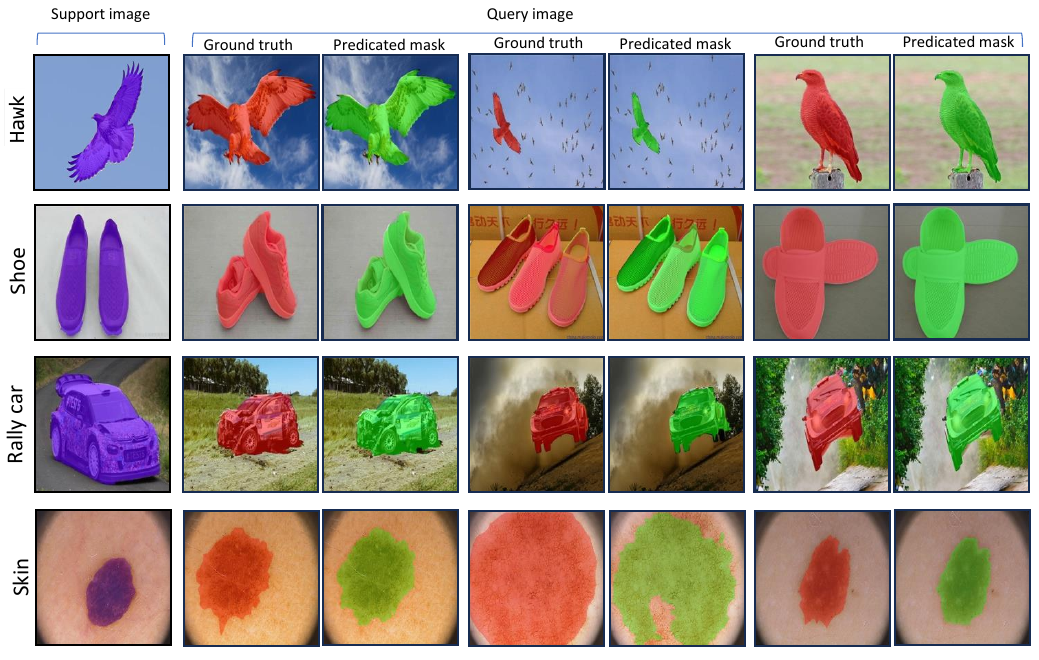}
    \caption{Sample of prediction results of the proposed method on the FSS-1000 and $PH^2$ datasets, employing a one-shot setting.}
    \label{fig:results}
\end{figure}

In the second setting, we conduct additional experiments without including support annotation. As outlined in our proposed method, we estimate the support distribution through spectral decomposition. Notably, our model performs exceptionally well even without annotation, as evident from \Cref{tab:results1}. In the 1-shot scenario, our model achieves a notable mIoU improvement of 19.7\% over the FSS-baseline method. In addition using the same setting, our method is abale to obtain superior performance than HSNet \cite{min2021hypercorrelation}. Some challenging cases are visualized in Figure~\ref{fig:results}.

\noindent\textbf{$PH^2$:} We present a comparative results in \Cref{tab:results2}. The comparative results highlight the superiority of our approaches over SOTA methods across all evaluation metrics, affirming the effectiveness of our self-supervised few-shot learning strategy. 
Notably, by employing the episodic training paradigm, a noticeable enhancement of approximately 19.1\% is observed compared to the few-shot baseline model suggested in~\cite{feyjie2020semi}. In contrast to the semi-supervised strategy~\cite{feyjie2020semi} that integrates additional samples through the utilization of unsupervised methodologies, the proposed models demonstrate a superior level of performance by employing a self-supervised strategy.
Moreover, our strategy differentiates itself from the self-supervised approach~\cite{ahn2021spatial,karimijafarbigloo2023ms} that generates a supervisory signal solely based on image content. Instead, we leverage a support sample to incorporate prior knowledge, effectively guiding the network and elevating its performance.
From a qualitative standpoint, we provide a visual comparison in Figure~\ref{fig:results}. 

\subsection{Ablation Study}
The proposed architecture incorporates two key modules: the CLKA, and the MS-AG module in the decoding path. These modules are designed to facilitate the feature representation and adaptively fuse support information into the query representation. In order to assess the impact and contribution of each module on the generalization performance, we conducted experiments where we selectively removed individual modules, as outlined in Table \ref{tab:modules}.
The experimental results highlight the significance of each module in the overall architecture. Specifically, removing any of the modules from the network leads to a noticeable decrease in performance. Notably, when the CLKA module is removed, the impact of support prior knowledege diminishes, resulting in a clear drop in performance. Similarly, replacing the MS-AG with simple concatenation results in a performance drop. However, by including the MS-AG module, our model tends to reduce the number of wrong predictions and isolated false positives.

\begin{table}[!htb]
	\centering
	\caption{(a) Comparison of IoU on the FSS-1000 dataset. (b) Comparative performance of the proposed method against the SOTA approaches on the PH$^2$ dataset. (c) Contribution of each module on the model performance.} \label{tab:main}
	\begin{subtable}[b]{0.54\textwidth}
		\caption{Results on FSS-1000 Dataset} \label{tab:results1}
        \centering
		\resizebox{\textwidth}{!}{
			\begin{tabular}{cl|cc}
				\toprule
				\multirow{2}{*}{\shortstack{Setting}} & \multirow{2}{*}{Methods} & \multicolumn{2}{c}{mIoU} \\ 
			
				& & 1-shot & 5-shot \\
				
				\midrule
				\multirow{2}{*}{With Annotation}      
                    & OSLSM~\cite{shaban2017one} & \underline{70.2} & \underline{73.0} \\
                    & co-FCN~\cite{rakelly2018conditional} & \underline{71.9} & \underline{74.2} \\
                    & FSS-1000~\cite{li2020fss} & \underline{73.4}& \underline{80.1} \\
                    & DoG~\cite{azad2021texture} & \underline{80.8} & \underline{83.3} \\
                    & PFENet~\cite{tian2020prior} & \underline{80.8} & \underline{81.4} \\
                    & MemoryFSS~\cite{lu2021learning} & \underline{83.0} & \underline{85.7} \\
                    & DAN~\cite{wang2020few} & \underline{85.2} & \underline{87.1} \\  & HSNet~\cite{min2021hypercorrelation} & \underline{85.5} & \underline{87.8} \\ \cline{2-4} \\[-2.0ex]
				\rowcolor[rgb]{0.682,0.859,0.855}
				& Proposed Method  & \textbf{85.7} & \textbf{87.9} \\
				\midrule
				\multirow{2}{*}{Without Annotation} & FSS baseline~\cite{feyjie2020semi} & \underline{65.3} & \underline{67.9} \\  & HSNet~\cite{min2021hypercorrelation} & \underline{84.3} & \underline{86.1} \\ \cline{2-4} \\[-2.0ex]
				\rowcolor[rgb]{0.682,0.859,0.855}
				& Proposed Method  & \textbf{85.0} & \textbf{86.8} \\				
				\bottomrule
			\end{tabular}
		}
	\end{subtable}
	\hfill
	\begin{subtable}[c]{0.40\textwidth}
		\caption{Results on PH2 dataset} \label{tab:results2}
        \centering
		\resizebox{\textwidth}{!}{
			\begin{tabular}{c||ccc} 
				\toprule
				\multirow{2}{*}{\textbf{Methods}} & \multicolumn{3}{c}{\textbf{PH$^2$}} \\ 
				\cline{2-4}
				& \textbf{DSC~$\uparrow$} & \textbf{HM~$\downarrow$} & \textbf{XOR~$\downarrow$} \\ 
				\hline
				FSS-baseline & 68.13 & - & - \\
				Semi-supervised FSS~\cite{feyjie2020semi} & 74.77 & - & - \\
				\textit{k}-means & 71.3 & 130.8 & 41.3 \\
				DeepCluster~\cite{caron2018deep} & 79.6 & 35.8 & 31.3 \\
				IIC~\cite{ji2019invariant} & 81.2 & 35.3 & 29.8\\
				SGSCN\cite{ahn2021spatial} & 83.4 & 32.3 & 28.2\\
				MSS-Former\cite{ahn2021spatial} & 86.0 & 23.1 & 25.9\\
				\hline
				\rowcolor[rgb]{0.682,0.859,0.855}
				\textbf{Our Method} & \textbf{87.3} & \textbf{21.2} & \textbf{23.5}\\
				\bottomrule
			\end{tabular}
		}
		\caption{Modules effect} \label{tab:c}
		\centering
			\resizebox{0.8\textwidth}{!}{
			\begin{tabular}{cc|c}
                \toprule
				\textbf{CLKA} & \textbf{MS{-}AG} & \textbf{mIoU (FSS-1000)} \\ 
				\midrule
				\xmark & \xmark & 83.8 \\
				\cmark & \xmark & 84.1 \\
				\xmark & \cmark & 84.0 \\
                    \rowcolor[rgb]{0.682,0.859,0.855}
				\cmark & \cmark & 85.0  \\
                \bottomrule
				\label{tab:modules}
			\end{tabular}
			}
	\end{subtable}
\end{table}

\section{Conclusion}
Our study presents a novel approach for addressing few-shot semantic segmentation on medical images in the absence of annotated data. We reframe the problem as a graph partitioning task and leverage the eigenvectors of the Laplacian matrix derived from self-supervised networks to effectively model the Support representation and capture the underlying distribution. Within the standard FSS framework, we predict the query mask by utilizing the learned support distribution. Furthermore, we introduce the hierarchical LKA module to enrich the feature representation and improve the decoding process.

\subsubsection{Acknowledgment}
This work was funded by the German Research Foundation (Deutsche Forschungsgemeinschaft, DFG)– project number 455548460.

\bibliographystyle{splncs04}
\bibliography{16.bib}

\end{document}